\documentclass[12pt,letterpaper]{article}
\usepackage[T1]{fontenc}
\usepackage[utf8]{inputenc}
\usepackage{lmodern} 
\usepackage{XCharter}
\usepackage[scaled]{helvet}
\usepackage{courier}
\usepackage{style/dsc180reportstyle} 
\usepackage[ruled,vlined,linesnumbered]{algorithm2e}
\usepackage{mathrsfs}
\usepackage{amsmath,amssymb}
\usepackage{booktabs} 
\usepackage{array}    
\usepackage{graphicx} 
\usepackage{tabularx} 
\usepackage{algpseudocode}
\usepackage{hyperref}

\title{FlyKD: Graph Knowledge Distillation on the Fly with Curriculum Learning}

\author{Eugene Ku \\
  {\tt euku@ucsd.edu} 
  \\}

\begin{document}
\maketitle



\begin{abstract}
\hspace{1cm}Knowledge Distillation (KD) aims to transfer a more capable teacher model's knowledge to a lighter student model in order to improve the efficiency of the model, making it faster and more deployable. However, the student model's optimization process over the noisy pseudo labels (generated by the teacher model) is tricky and the amount of pseudo labels one can generate is limited due to Out of Memory (OOM) error. In this paper, we propose FlyKD (Knowledge Distillation on the Fly) which enables the generation of virtually unlimited number of pseudo labels, coupled with Curriculum Learning that greatly alleviates the optimization process over the noisy pseudo labels. Empirically, we observe that FlyKD outperforms vanilla KD and the renown Local Structure Preserving Graph Convolutional Network (LSPGCN). Lastly, with the success of Curriculum Learning, we shed light on a new research direction of improving optimization over noisy pseudo labels. 
\end{abstract}

\begin{center}
Website: \url{https://drug-repurposing-gnn.github.io/Drug-Repurposing-Website/} \\
Code: \url{https://github.com/Drug-Repurposing-GNN/SSL-DiseaseDrug-Prediction}
\end{center}

\maketoc
\clearpage


\section{Introduction}
\hspace{1cm}Graph Neural Networks (GNN) are successful and powerful tools for recognizing complex patterns within graphs while utilizing its topological structure. Tremendous success in applications of GNNs are shown in Antibiotic Discoveries [\cite{stokes2020deep}], Recommendation Systems [\cite{RecommendationSystems}], and Nutrition Services [\cite{tian2022reciperec}], to name a few. 

\hspace{1cm} However, despite being powerful and often computationally efficient, GNNs are not memory friendly [\cite{Eugenespaper}], which makes it cumbersome to deploy (e.g. to users devices or as part of a product). Thus, a form of model compression is essential for the deployment of GNNs. 

\hspace{1cm} Among many forms of model compression, Knowledge Distillation (KD) is a popular choice [\cite{geoffhintonKD, GNNKD}]. Generally, Knowledge Distillation involves a teacher model and a student model where the more capable teacher model generates pseudo labels. These pseudo labels are then utilized by the student model to augment its training process. Due to the additional optimization process in Knowledge Distillation (i.e. training of a teacher model), the pseudo labels are inherently noisy and difficult to optimize on. 

\hspace{1cm} Despite this issue, the current state of research in Knowledge Distillation is often only concerned with how to generate the pseudo labels (What to distill) but completely neglects how to stably train the student model on inherently noisy pseudo labels. 

\hspace{1cm} To address this issue, we incorporate a form of Curriculum Learning [\cite{curriculum}]. Inspired by how humans learn, Curriculum Learning is an optimization method that allows models to achieve a better performance by first introducing easy, simple labels then gradually increasing the difficulty by introducing labels that rely on more complex patterns. 

\hspace{1cm} To incorporate Curriculum Learning, we use our prior knowledge of noisiness of each label to base the difficulty, then use the inferred difficulty to facilitate the order of labels the student model should see. We show that this guidance of optimization leads to substantial improvement where negative Knowledge Distillation can become positive (Table \ref{tab:1}). 

\hspace{1cm} Furthermore, with our successful incorporation of Curriculum Learning, stabilizing the training process over noisy pseudo labels, we also propose a new KD method that focuses on the quantity of the pseudo labels over quality. 

\hspace{1cm} We propose FlyKD, a noisy Graph Knowledge Distillation framework for link prediction task that can generate a stupendous amount of pseudo labels on the fly while avoiding out of memory (OOM) error. We show that by storing the probability scores on the links of newly generated random graph per epoch, we can generate 100-1000x or more pseudo labels beyond the threshold of the traditional KD methods. Under both no Curriculum-Learning and with Curriculum-Learning setting, we observe that FlyKD shows noticeable improvement over the vanilla KD and LSPGCN [\cite{distilgcn}] to achieve the state-of-the-art performance.

\section{Related Works}
\subsection{Graph Neural Networks}

\textbf{Notations}. We denote {$\mathbf{\mathcal{G}}$ = ($\mathbf{\mathcal{V}, \mathcal{E}, X}$)} as a graph, where {$\mathbf{\mathcal{V} = \{v_1, v_2, ..., v_N\}}$} represents the set of nodes, {$\mathbf{\mathcal{E} \subseteq \mathcal{V} \times \mathcal{V}}$} represents set of edges, {$\mathbf{X}\in\mathbb{R}^{N \times F}$} is the node feature matrix, and {$\mathbf{\bigoplus}$} represents any size-invariant aggregation function such as \textbf{min, max, sum, average}, whereas \textbf{AGG} can be any aggregators including size-variant ones such as concatenation. Furthermore, $\mathbf{h^{(l)}}$ denotes the hidden node features at layer $l$, while $\mathbf{\mathcal{N}}$ denotes the 1-hop neighbors of a particular node. 

\textbf{Message Passing Neural Networks}. Graph Neural Networks (GNN) are Neural Networks [Deep Learning] designed to achieve permutation-invariance and size-invariance on graphs, where no matter the order of node id, we can obtain the same output (permutation-invariance) and is also generalizable to graphs of varying sizes (size-invariance).

\hspace{1cm}Among all, the most popular framework of GNN is the Message Passing Neural Network (MPNN). In the MPNN framework, nodes exchange messages with its nearby neighbors, thereby incorporating the topological structure of the graph in their learning. The MPNN framework for \textbf{node-level task} can be formulated as follows: 
\begin{equation}
\label{eq:1}
\forall v \in \mathcal{V}, \hspace{5pt} h^{(l)}_v := \text{AGG}(\phi(h^{l-1}_v), \bigoplus_{\substack{u \in \mathcal{N}(v)}} \psi(h^{l-1}_u))
\end{equation}

where $\phi$ and $\psi$ can be any sub-differentiable function such as Multi-Layer Perceptron [MLP].



\hspace{1cm}Popular GNN architecture that uses MPNN framework include Graph Convolutional Network (GCN), Graph Attention Network (GAT), Graph Isomorphic Network (GIN) [\cite{GCN, GAT, GATv2, GIN}].

\textbf{Knowledge Graph Embedding.}
First popularized by Google in 2012 [\cite{Google2012KnowledgeGraph}], Knowledge Graphs (KG) are often structured to encapsulate a comprehensive knowledge in the form of massive number of entities and relation between entities. Such KG can be used for machine learning tasks such as completing the missing relations between nodes given the known relations. 

\hspace{1cm}However, standard GNN framework alone is not sufficient for such task due to the sheer size of KG and common absence node/edge features other than the node types, edge types, and the adjacency matrix.  

\hspace{1cm}Thus, to scalably generate meaningful embeddings for nodes and edges, efficient scoring functions were created to complement the existing GNN framework. By and large, these scoring functions for KG can be categorized into two categories: distance-based vs. semantic-based. 
In our baseline, we incorporate a semantic-based scoring function, DistMult [\cite{distmult}] which can be formulated as follows:

\begin{equation}
\label{eq:2}
\textbf{DistMult}(h, r, t) = \sum_{i=1}^{d}t_i * r_i * h_i
\end{equation}
where $h$ (head) represents source node embedding, $r$ represents relation embedding, and $t$ (tail) represents the target node embedding. After applying the scoring function, one can easily apply a Softmax or a Sigmoid function to obtain the probability score of any link existing between two nodes. 

\hspace{1cm}Lastly, due to the heterogeneous nature (containing many node types) of Knowledge Graphs, Relational GNN (\textbf{R-GNN}) [\cite{relational}] is often employed where the transformation function $\psi_r$ (applied on the messages) depend on the relation and can be formulated as follows:
\begin{equation}
\forall v \in \mathcal{V}, \hspace{5pt} h^{(l)}_v := \text{AGG}(\phi(h^{l-1}_v), \bigoplus_{r \in R}\bigoplus_{\substack{u \in \mathcal{N}_r(v)}} \psi_{r}(h^{l-1}_u))
\label{eq:3}
\end{equation}
where R represents the set of relations. Note that this is identical to equation \ref{eq:1} except that $\phi$ depends on r and the relation-specific messages are aggregated once more across relations. 

\subsection{Knowledge Distillation on Graphs}
\hspace{1cm}Knowledge Distillation generally involves a teacher and student model, where the goal of the teacher model is to transfer its knowledge to the student model. In order to allow such learning, \textbf{logits, intermediate-layer features, activation of neurons, or other representational embeddings} can all be used as \textbf{self-supervised labels} for the student model [\cite{geoffhintonKD}, \cite{heoetal.}]. A divergence loss between teacher and student model is as follows: 
\begin{equation}
\mathbf{\mathcal{L}_{DIV} = DIV(k^T, k^S)}
\end{equation}
where DIV stands for divergence loss (e.g. Kull-Back Divergence Loss), $k^T$ and $k^S$ stand for knowledge obtained by teacher and student model, respectively. Depending on the architecture, the knowledge $k$ is replaced by any or all of aforementioned self-supervised labels. $\mathcal{L}_{KD}$ is then summed up with the original loss from the original dataset to jointly train and looks as follows:
\begin{equation}
\mathbf{\mathcal{L}_{KD} = \lambda*\mathcal{L}_{org} + \mathcal{L}_{DIV}}
\end{equation}
where KD stands for Knowledge Distillation and org stands for original. Here, $\mathcal{L}_{org}$ facilitates the learning from the original dataset while $\mathcal{L}_{DIV}$ facilitates the learning from the pseudo labels generated by the teacher model. Intuitively, pseudo labels contain more information than simple ground truth label since it contains how confident the model should be and any similarity across classes in non-binary classification tasks. 

\hspace{1cm}Furthermore, one can intuitively view the utilization of the ground truth label as to encourage the student model to err towards the ground truth label when it is unable to fully imitate the behaviour of the teacher model. Thus, $\lambda$ is often small (<$0.05$) to put more emphasis on the pseudo labels. 

\subsection{PrimeKG and TxGNN}
\label{sec:2.3}

\begin{figure}
    \centering
    \includegraphics[width=0.5\linewidth]{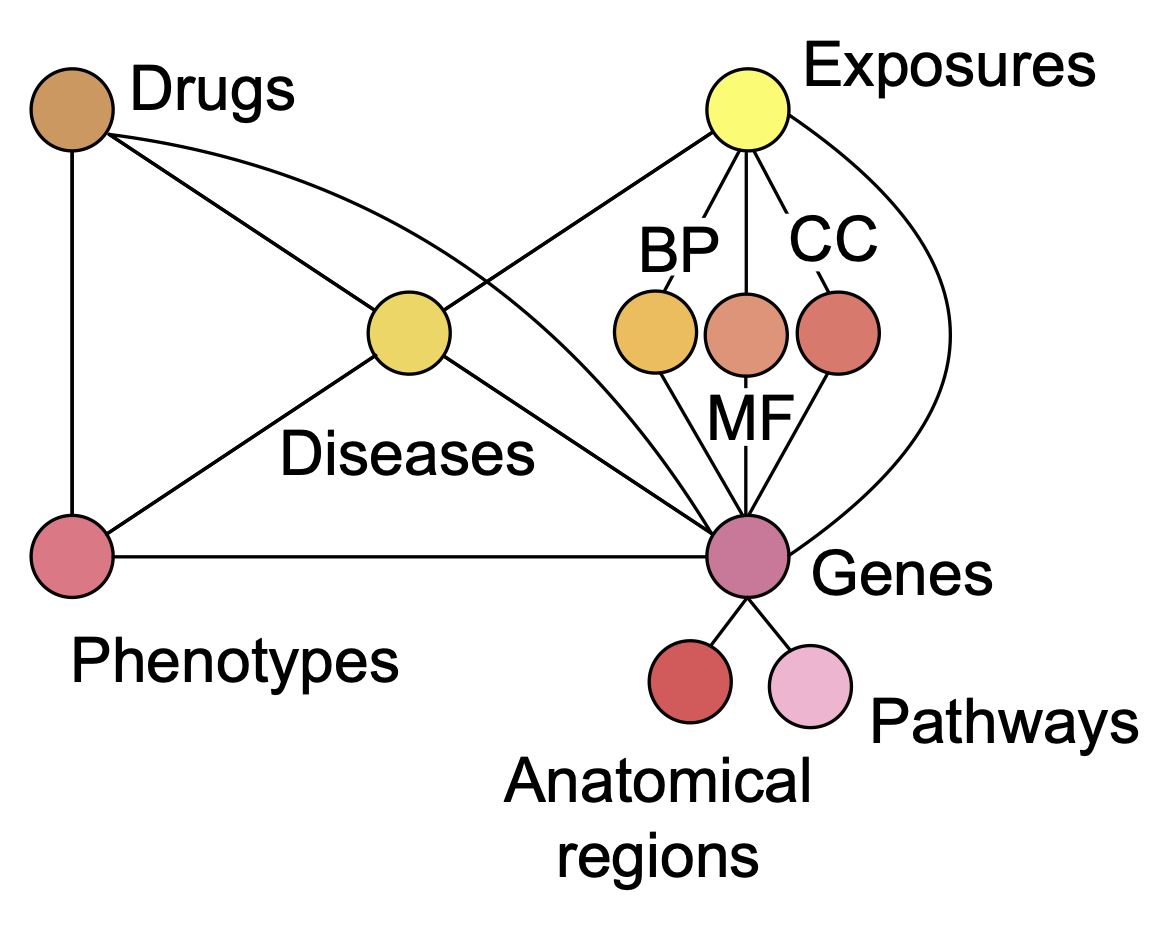}
    \caption{\textbf{Illustrative representation of PrimeKG dataset.} PrimeKG is a collaborative dataset, scraped from 20+ high quality databases. Adapted from [\cite{PrimeKG}]}
    \label{fig:PrimeKG}
\end{figure}

\textbf{PrimeKG.} Precision Medical Knowledge Graph (PrimeKG) [\cite{PrimeKG}] is a biomedical Knowledge Graph that consolidates a wide array of biomedical research information into a unified graph with new insights into biological processes and diseases. It is a critical tool for researchers in genomics, pharmacology, and related fields where they can explore biological networks and identify potential therapeutic targets. In particular, PrimeKG features 10 unique node types and 29 types of edges, encapsulating over 4 million relations and is illustrated in Figure \ref{fig:PrimeKG}.

\textbf{TxGNN.} TxGNN [\cite{TxGNN}] is a baseline model that will be used throughout this paper to benchmark our proposed methods. Specifically, TxGNN is a variant of R-GCN (Eq. \ref{eq:3}) that is composed of two phases: pretraining and finetuning. During pretraining, TxGNN aims to predict all the relations within the KG but aims to only predict a curated set of relations during the finetuning phase. 

\hspace{1cm}Furthermore, TxGNN is specially designed for zero-shot prediction, making it particularly suitable for drug repurposing and the identification of potential treatments for diseases with no existing medicines. TxGNN implements a novel disease pooling mechanism to excel at zero-shot settings, which pulls well-informed diseases' embeddings to augment the poorly-informed diseases that lacks drug-disease relations by computing the similarity between diseases. Since poor disease embeddings are unreliable, we assign a one-hop neighborhood binary vector [\cite{biologynetwork}] to each disease to compute the similarity score. 

\section{Methods}
\subsection{FlyKD}
\hspace{1cm}As the name suggests, FlyKD generates pseudo labels on the fly as opposed to traditionally pre-computing the pseudo labels before training. Generating pseudo labels on the fly allows FlyKD to go beyond the limitation of GPU memory and thus, allows the student model to get exposed to more diverse behaviour of the teacher model.

\hspace{1cm} Especially, FlyKD generates diverse set of pseudo labels on the fly by computing the probability score on the links of a random graph generated at each epoch. To maintain the topological structure of the original graph, we first obtain the final node embeddings by running GNN on the original graph, then apply DistMult (Eq. \ref{eq:2}) scoring function on the random graph to obtain the final predicted probability on each links. Intuitively this allows the student model to imitate the teacher model at a global level as links can form between any two node-pair in the random graph. Specifically, the algorithm for generating the random graph is formulated in Algorithm \ref{alg:1}.

\begin{algorithm}[H]
\label{alg:1}
\caption{GenerateRandomGraph (Degree-Aware)}
\DontPrintSemicolon  
\KwIn{Set of nodes $\mathcal{V}$, Set of relations $\mathcal{R}$, Relation-specific node degree matrix $\mathcal{D} \in \mathbb{N}^{|\mathcal{R}| \times |\mathcal{V}|}$, Number of pseudo labels to generate $k$.}
\KwOut{Set of edges for the random graph $\mathcal{E}_r \subseteq \mathcal{R} \times \mathcal{V} \times \mathcal{V}$.}
Initialize $\text{src}$ and $\text{dst}$ as empty lists for each relation in $\mathcal{R}$.\;
\For{rel in $\mathcal{R}$}{
    $\text{src}_{\text{rel}} \gets \text{Multinomial}(k, \text{weights}=\mathcal{D}_{\text{rel}})$\;
    $\text{dst}_{\text{rel}} \gets \text{Multinomial}(k, \text{weights}=\mathcal{D}_{\text{rel}})$\;
}
$\mathcal{E}_r \gets \text{stack}(\text{src}, \text{dst})$\;
\Return{$\mathcal{E}_r$}\;
\end{algorithm}

\hspace{1cm}In plain words, GenerateRandomGraph selects two nodes where the probability of a node being selected is proportionate to the degree of a node in the original graph (with respect to the relation). By doing so, we increase the quality of the pseudo labels on the random graph by utilizing the prior information that nodes that have seen more labels during training are more likely to have a higher embedding quality. Once two nodes are selected, a link is formed between them. This process of generating pseudo links is repeated k times. 

\begin{figure}
    \centering
    \includegraphics[width=.75\linewidth]{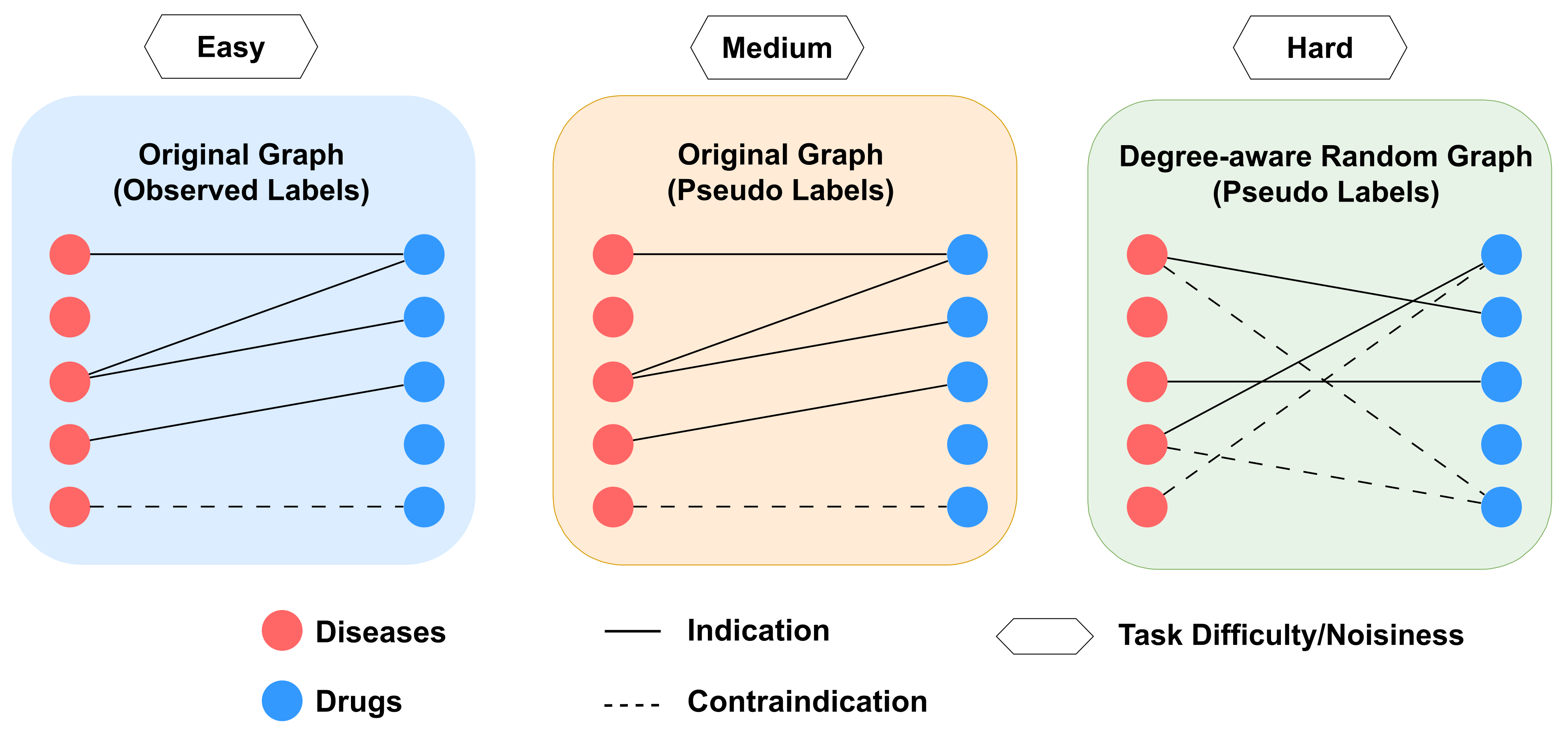}
    \caption{\textbf{Three types of labels in FlyKD.} Note that without the pseudo labels on the Degree-aware Random Graph (Green), FlyKD is equivalent to vanilla Knowledge Distillation.}
    \label{fig:2}
\end{figure}

\hspace{1cm} With the addition of pseudo labels on the fly, FlyKD has a total of three types of labels: original label, pseudo label on original graph, and pseudo labels on the random graph as illustrated in Figure \ref{fig:2}. The balance of these three types of labels are then controlled by Curriculum Learning (Section \ref{sec:3.2}) to further address the noisiness and randomness of pseudo labels. Overall, the entire architecture of FlyKD is formulated in Algorithm \ref{alg:2}.

\begin{algorithm}[H]
\label{alg:2}
\caption{FlyKD}
\DontPrintSemicolon
\KwIn{Train graph $G_t$, Observed labels $Y_o$, Number of pseudo labels to generate $k$, LossWeightScheduler, Teacher model $T$, Student model $S$}
\KwOut{Trained student model with Teacher's Knowledge $S_{KD}$}
Train $T$ with $Y_o$\;
$h^T \gets T(G_t)$ \tcp*{Obtain Final Node Embedding (Teacher)}
$Y_{pt} \gets \text{DistMult}(h^T, G_t)$ \tcp*{Generate pseudo labels}
\For{each epoch $e$}{
    $h^S \gets S(G_t)$ \tcp*{Obtain Final Node Embeddings (Student)}
    $G_r \gets \text{GenerateRandomGraph}(G_t, K)$\;
    $Y_r \gets \text{DistMult}(h^T, G_r)$\tcp*{Generate Pseudo Labels (Random graph)}
    $\hat{Y_t} \gets \text{DistMult}(h^S, G_t)$ \tcp*{Predict on Train Graph}
    $\hat{Y_r} \gets \text{DistMult}(h^S, G_r)$\tcp*{Predict on Random Graph}
    $\mathcal{L}_{og} \gets \text{Loss}(\hat{Y_t}, Y_o)$\tcp*{Loss from Original Label}
    $\mathcal{L}_{pe} \gets \text{Loss}(\hat{Y_t}, Y_{pt})$\tcp*{Loss from Pseudo Label (Train Graph) }
    $\mathcal{L}_{pr} \gets \text{Loss}(\hat{Y_r}, Y_r)$\tcp*{Loss from Pseudo Label (Random Graph)}
    $\lambda_{og}^{(e)}$, $\lambda_{pe}^{(e)}$, $\lambda_{pr}^{(e)} \gets \text{LossWeightScheduler}(e)$\;
    $\mathcal{L}_{total} \gets \lambda_{og}^{(e)} * \mathcal{L}_{og} + \lambda_{pe}^{(e)} * \mathcal{L}_{pe} + \lambda_{pr}^{(e)} * \mathcal{L}_{pr}$\;
    GradientDescent(S, $\mathcal{L}_{total}$) \tcp*{Update model parameters}
}
return S
\end{algorithm}

\subsection{Curriculum Learning}
\label{sec:3.2}
\begin{figure}
    \centering
    \includegraphics[width=0.75\linewidth]{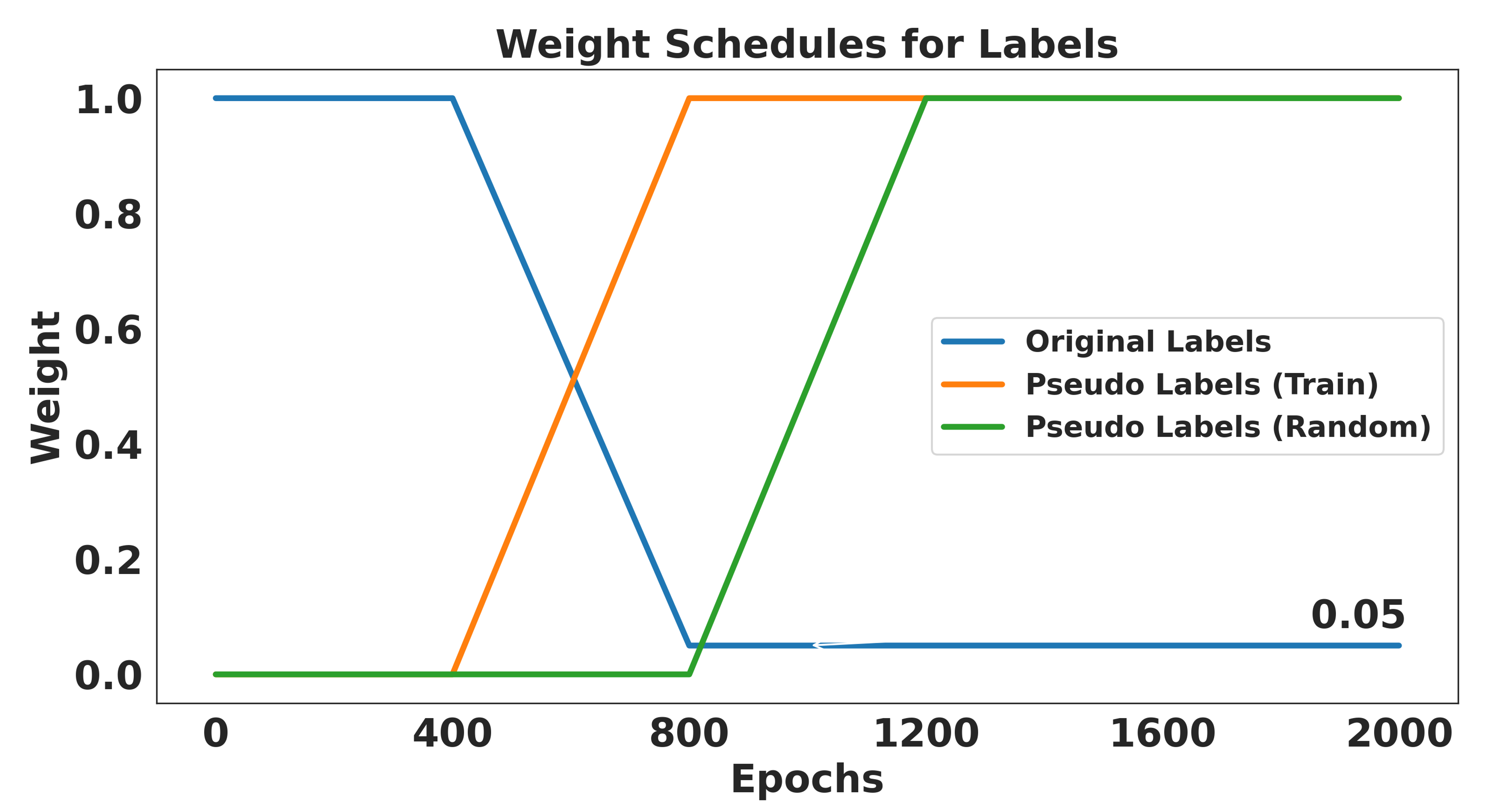}
    \caption{\textbf{Plot of loss scheduler that incorporates Curriculum Learning.} The colors of loss schedulers match with that of Figure \ref{fig:2}. Notice that the original loss doesn't completely go to 0 but 0.05 to avoid catastrophic forgetting.}
    \label{fig:3}
\end{figure}

\hspace{1cm}Generating tremendous amount of pseudo labels comes at a cost: the pseudo labels are extra noisy. To combat this issue, we employ Curriculum Learning which is by and large viewed in two categories based on how one measures the difficulty of each data:
\begin{enumerate}
    \item Use prior knowledge of the data
    \item Create a custom difficulty metric (e.g. entropy score) [\cite{curriculumlearningGNN}]
\end{enumerate}
\hspace{1cm}We use the former as we have a good idea on which labels are more noisy thus more difficult to optimize. To enforce a gradual transition of emphasis on three losses, we incorporate a linear loss scheduler as depicted in Figure \ref{fig:3}. 

\hspace{1cm}One can analytically view our approach of first utilizing less-noisy ground truth as optimizing over a smooth function first then gradually transitioning into more complex, noisy labels that corresponds to a non-smooth function. This allows FlyKD to warm-up and gain a good initialization point in the beginning which is paramount for optimal optimization in stochastic optimization [\cite{weight1, weight2}].




\section{Results}
\hspace{1cm} Out of all the relations in PrimeKG, we evaluate on predicting indication and contraindication relations with macro AUPRC as our evaluation metric. We specifically focus on these relations as they directly relate to the performance of drug repurposing task. For the baseline architecture, we incorporate aforementioned TxGNN (section \ref{sec:2.3}) for both our teacher and student model, where the teacher model has an embedding size of 130 while the student model has 80. The goal for the evaluated approaches is to distill teacher model (130) to student model (80) where the distilled student model (80) outperforms the baseline model (80) that has not received any KD. The performance of baseline models are listed in Table \ref{tab:1}.
\newline
\newline

\begin{table}[ht]
\caption{\textbf{Teacher and Student Models' Performances.} The AUPRC is scaled by 100 and averaged by macro. Baseline 130 is the Teacher model and Baseline 80 is the Student Model.}
\label{tab:1}
\centering
\resizebox{\textwidth}{!}{%
\begin{tabular}{@{}cccccccc@{}}
\toprule
\multicolumn{8}{c}{Baseline AUPRC (\%)} \\
\midrule
Model & Num. Params & Seed 45 & Seed 46 & Seed 47 & Seed 48 & Seed 49 & Mean $\pm$ std \\
\midrule
Baseline 130 & (1.7M)	& 80.33	& 73.66	& 76.29	& 84.19	& 79.91	& 78.87 $\pm$ 4.04 \\
Baseline 80 & (650k)	& 78.64	& 71.97	& 74.44	& 82.74	& 77.87	& 77.13 $\pm$ 4.13 \\
\end{tabular}%
}
\end{table}

\hspace{1cm}Due to the nature of zero-shot evaluation setting, the standard deviation across different seeds is high. Thus, we compute the relative performance difference from incorporating a method per seed and then average across seed. This way, we can more accurately attribute changes in performance to the methods themselves, rather than to fluctuations in task difficulty associated with different seeds. 

\hspace{1cm}We mainly explore three KD method for GNN: Basic Knowledge Distillation (BKD) [\cite{geoffhintonKD}], Local Structure Preserving GCN (LSPGCN, aka. DistillGCN) [\cite{distilgcn}], and finally our proposed method FlyKD. 

\hspace{1cm}We observe that both BKD and LSPGCN show negative relative performance difference while only FlyKD shows positive gains over the baseline model (80), which did not receive any KD  (Table \ref{tab:2}). From our ablation study, we find out the reason for such gap between our FlyKD and other KD methods is due to the noisiness of teacher-generated pseudo labels. When we employed Curriculum Learning to BKD, we observe a dramatic boost in performance (+1.55\%) compared to without as shown in Table \ref{tab:3}.
\begin{table}[ht]
\caption{\textbf{KD Showdown.} Only FlyKD incorporates Curriculum Learning and thus, shows positive gains.}
\label{tab:2}
\centering
{\small 
\begin{tabular}{@{}cccccccc@{}}
\toprule
\multicolumn{4}{c}{Knowledge Distillation Methods (Relative gains from Baseline80)} \\
\midrule
Model & Time & Curriculum Learning & Mean$\pm$std \\
\midrule
Basic KD & 1600 & No &  -0.62$\pm$0.59 \\
LSP 1 layer (RBF) & 20000 & No & -1.09$\pm$0.23 \\
LSP 2 layers (RBF) & 40000 & No & -1.41$\pm$0.82 \\
FlyKD & 2000 & Yes & \textbf{1.16$\pm$0.36} \\
\end{tabular}
}
\end{table}
{
\begin{table}[ht]
\caption{\textbf{Ablation Study.} Curriculum Learning seems to play a pivotal role.} 
\label{tab:3}
\centering
{\small
\begin{tabular}{@{}ccc@{}}
\toprule
\multicolumn{3}{c}{Ablation study (Relative gains from Baseline80)} \\
\midrule
Model & Configuration & Mean$\pm$std \\
\midrule
Basic KD & Employ Curriculum Learning & 0.93$\pm$0.45 \\
FlyKD & Fix Random Graph & 1.14$\pm$0.39 \\
FlyKD & No Curriculum Learning & 0.19$\pm$0.42 \\
FlyKD & Take Out Pseudo Labels on Train Dataset & -0.68$\pm$0.63 \\
FlyKD & stepwise function for Curriculum Learning & -1.436$\pm$0.86 \\
\end{tabular}
}
\end{table}}

\hspace{1cm}We also consider a step-wise curriculum learning where the transition between different phases of the loss scheduler is instant. However, we observe a major degrade in performance (Table \ref{tab:3}), emphasizing the importance of gradual change in difficulty in Curriculum Learning. 

\section{Discussion}
\textbf{Future Works:} Despite FlyKD's success in PrimeKG, we admit that experiments with one dataset gives limited validations for our findings. Thus, more experiments with diverse dataset is direly needed. 

\textbf{Findings.} We briefly sum up our findings here.
\begin{enumerate}
    \item Optimization process of the student model in KD is inherently noisy and Curriculum Learning can greatly alleviate this issue. 
    \item  Curriculum Learning for KD needs gradual change in difficulty, corroborated by catastrophic degrade in performance when step-wise loss scheduler is employed instead. 
    \item LSPGCN is not scalable for Knowledge Graphs, demonstrated by 10-20x increase in time as seen in Table \ref{tab:2}. This holds true for most Knowledge Distillation tailored for graphs with similar mechanism. 
    \item FlyKD's pseudo labels on degree-aware random graphs seem to help but doesn't provide any more gains from sticking with one random graph instead of generating new random graph at every epoch throughout the training. Further investigation is needed to understand this phenomenon. 
\end{enumerate}




\section{Conclusion}
\hspace{1cm}In our paper, we propose a novel KD method called FlyKD which enables a generation of virtually unlimited amount of pseudo labels without running into any memory errors. We also address the weakness of FlyKD (noisiness of pseudo labels on random graph) by incorporating Curriculum Learning. Surprisingly, the Curriculum Learning seems to also drastically improve vanilla KD too, suggesting a new research direction of how to improve the optimization process of the student model rather than the common what pseudo labels to generate (What to distill).


\makereference

\bibliography{paper}
\bibliographystyle{style/dsc180bibstyle}


\clearpage
\makeappendix

\subsection{Training Details}
\hspace{1cm}We use Adam [\cite{adam}] as our optimizer with the learning rate (lr) of 0.001 during pretraing phase for 1 epoch while we use the lr of 0.0005 during the finetuning phase for 2000 epochs. We observed a major decline in performance when pretraining for longer than 1 epoch. For the finetuning phase, we incorporated a Reduce Learning Rate on Pleatue by 0.8 for the last 400 epochs to help the convergence.

\hspace{1cm}Further, Negative sampling was used on the original graph to avoid only predicting positive labels. We matched the number of negative labels to the number of positive labels in the original graph. Since we are using AUPRC as our evaluation metric, the performance is robust to the cut-off threshold and thus, the ratio of positive and negative labels did not matter.

\subsection{Additional Experiments}
Here, we leave three other variations of FlyKD that we tried but did not come to fruition. Strong Scores (2) modification refers to only maintaing pseudo labels with high confidence (>logit score of 2). Occasional Random Graph (5) refers to only switching to a newly generated Random Graph at every 5 epoch. Mod probability refers to modifying the probability distribution of a random graph where we decrease the entropy of it by raising it to the probability by 1.5 and re-normalizing. All three variations either did worse or matched the relative performance gains of the original FlyKD.
\begin{table}[ht]
\label{tab:4}
\centering
{\small 
\begin{tabular}{@{}cc@{}}
\toprule
\multicolumn{2}{c}{Additional Knowledge Distillation Methods (Relative gains from Baseline80)} \\
\midrule
Configuration (FlyKD) & Mean$\pm$std \\
\midrule
Strong Scores (2) & 0.02$\pm$0.23 \\
Occasional Random Graph (5) & 1.16$\pm$0.35 \\
Mod Probability & 1.15$\pm$0.36 \\
\end{tabular}
}
\end{table}

\subsection{Acknowledgements}
Finally, we would like to especially thank Gal and Yusu for their guidance on our project for many months. From offering many meetings during summer and throughout two quarters to answering our persistent questions through email and slack, our gratitude exceeds the capability of expressing it.

\end{document}